\documentclass[fullname,11pt]{article}
\usepackage[T1]{tipa}
\usepackage{fullpage,epsfig,fullname}

\newcommand{\xng}{\textipa{\ng}}
\newcommand{\xsh}{\textipa{\textesh}}
\newcommand{\xch}{\textipa{\textteshlig}}
\newcommand{\xdzh}{\textipa{\textdyoghlig}}
\newcommand{\xdh}{\textipa{\dh}}
\newcommand{\xth}{\textipa{\texttheta}}
\newcommand{\xzh}{\textipa{\textyogh}}

\newcommand{\xI}{\textipa{I}}

\newcommand{\xae}{\textipa{\ae}}
\newcommand{\retfric}{\textipa{\textrtails}}
\newcommand{\alvpalfric}{\textipa{C}}   
\newcommand{\xrUS}{\textipa{\textturnr}}   	
\newcommand{\Gr}{\textipa{\textinvscr}}		
\newcommand{\Dg}{\textipa{\textgamma}}		
\newcommand{\Gch}{\textipa{\c{c}}}		
\newcommand{\Hat}[1]{\stackrel{\wedge}{{#1}}}
\newcommand{\FL}{\Hat{F\hspace{-0.5ex}L}}
\newcommand{\HH}{\Hat{H}\hspace{-0.5ex}}

\title{Measuring the Functional Load of Phonological Contrasts}

\author{Dinoj Surendran\footnote{University of Chicago, Department of Computer Science, dinoj@cs.uchicago.edu},
 Partha Niyogi\footnote{University of Chicago, Departments of Computer Science and Statistics, niyogi@cs.uchicago.edu}}

\begin{document}

\maketitle

\textit{ Frequency counts are a measure of how much use a language
makes of a linguistic unit, such as a phoneme or word. However, what
is often important is not the units themselves, but the contrasts
between them. A measure is therefore needed for how much use a
language makes of a contrast, i.e. the functional load (FL) of the
contrast.  We generalize previous work in linguistics and speech
recognition and propose a family of measures for the FL of several
phonological contrasts, including phonemic oppositions, distinctive features,
suprasegmentals, and phonological rules. We then
test it for robustness to changes of corpora. Finally, we provide
examples in Cantonese, Dutch, English, German and Mandarin, in the
context of historical linguistics, language acquisition and speech
recognition.  More information can be found at
http://dinoj.info/research/fload .  }

\tableofcontents

\section{Introduction}

\begin{quote}
``The term functional load is customarily used in linguistics to
describe the extent and degree of contrast between linguistic units,
usually phonemes.  In its simplest expression, functional load is a
measure of the number of minimal pairs which can be found for a given
opposition. More generally, in phonology, it is a measure of the work
which two phonemes (or a distinctive feature) do in keeping utterances
apart -- in other words, a gauge of the frequency with which two
phonemes contrast in all possible environments'' -- \namecite{king}
\end{quote}

This paper describes a method to measure how much use a language makes
of a contrast to convey information, i.e. the functional load (FL) of
the contrast.

The concept of FL goes back to the 1930s. However, existing
definitions are so limited that researchers who want to measure FL
often cannot. For example, \namecite{pye} speak of the need to make
explicit a phonological model of acquisition which ``predicts that
children will attempt to build phonemic contrasts on the basis of
maximal opposition within the language''. They go on to say :

\begin{quote}
``We need a rigorous definition of maximal oppositions that specifies
the relative strengths of different features within any
language. $\ldots$ The frequency of consonants across lexical types is
an imperfect guide to children's phonological systems because it
refers to isolated segments rather than oppositions.'' --- \namecite{pye}
\end{quote}

\namecite{ingram} suggests a method of computing FL, based on counts of
minimal pairs, but as \namecite{sododd} point out, it ``does not
include other aspects of phonology that might contribute, relatively,
to the functional loading of consonants: vowel, syllable structure,
stress and tone.''

The framework we propose {\em does} measure the FL of consonant
oppositions, and several other contrasts, while taking into
consideration word and syllable structure, stress and tone.  The use
of the term `contrast' in this paper is broader than standard,
encompassing phoneme oppositions (binary or not), distinctive features
(again, binary or not), suprasegmental features and even phonological
rules such as phoneme deletion in certain contexts. This permits
researchers with the appropriate corpora to answer questions like
these:

\begin{itemize}
\item Is it more important to correctly hear the tone or the vowel in
Cantonese?
\item Does Hindi make more use of aspiration or voicing?
\item How much information is lost due to vowel reduction in
unstressed syllables?
\item If second-language speakers have trouble
 learning contrasts that are not present in their native language,
 e.g. the [l]-[r] distinction in English for Japanese speakers, 
 how badly off are they?
\end{itemize}.


Section \ref{history} summarizes the history of FL in linguistics and
related work in speech recognition. Section \ref{FLdef} defines our FL
measure and Sections \ref{egtypes} and \ref{egcontrasts} demonstrate
the range of its applicability with several examples. Section
\ref{testconsistency} tests its robustness to the approximations
required to compute it. Section \ref{nonideal} is similar,
investigating whether corpora that are not representative of
continuous speech, such as word-frequency lists with citation form
pronunciations and written frequencies, give usable FL values.
Sections \ref{difflang} to \ref{applic_ASR} give detailed examples of
applications in linguistic typology, historical linguistics, language
acquisition and speech recognition. Applications come with actual
computations with corpora for Cantonese, Dutch, English, German and
Mandarin. Section \ref{interpret} discusses the interpretation of FL
values, especially in light of their being relative values rather than
absolute.

As we have not managed to eliminate enough notation from them, readers
may wish to skim Section \ref{FLdef} and skip Section
\ref{testconsistency} on a first reading.

\section{Previous Work}\label{history}
 
\subsection{FL in the Linguistics community}

Languages use contrasts of features to convey information.
The concept of `amount of use a language makes of a contrast' arose in
linguistics early in the 20th century, and the term \textit{functional
load} for a measure of it can be found in the writings of the Prague School
\cite{mathesius,trubetzkoy}.  The term `contrast' was nearly always
taken to mean `binary opposition of phonemes'. 

\namecite{martinet} popularized the concept, positing it as an
important factor in sound change. This has been disputed; a
quantitative corpus-using study by \namecite{king} found no evidence
for FL playing a role in the context of phonological mergers. But
finding no evidence for X and finding evidence against X are 
different things, and the reader interested in the debate is
referred to Peeters \shortcite{peeters},  Lass
\shortcite{lass80,lass97}, and to the example in the case of a recent
merger in Cantonese in Section \ref{applic_histling}. 

\namecite{meyerstein} notes, in his survey of the topic, that 
FL is easy to define intuitively but hard to define precisely. The
first person to propose a formula for it was \namecite{hockett}.  His
formula was only meant for the FL of the opposition of a pair of
phonemes, say $x$ and $y$, in a language $L$\footnote{\namecite{wang}
generalized Hockett's definition to the opposition between elements of
a set of phonemes.}. The absence of this opposition
creates a language $L_{xy}$ just like $L$ but with $x$ and $y$
collapsed into a single phoneme.  For example, in $English_{bp}$ the
verbs `bat' and `pat' have the same pronunciation.

Hockett assumed that any language could be modelled by a sequence of
phonemes, and its informational content represented by the entropy $H$
of a language. (The definition and computational details of $H$ are
described in Section \ref{FLdef}. For now, we just need to know that
$H$ is the number of bits of information transmitted by the language.)
The closer $H(L)$ and $H(L_{xy})$ are, the less the information lost when
the $x-y$ opposition is lost from $L$, and hence the less the reliance
of $L$ on it. Therefore he proposed :

\begin{equation}\label{FLhockett}
FL_{Hockett}(x,y) = \frac {H(L) - H(L_{xy})}{H(L)}
\end{equation}

The crucial part of the definition is the numerator, which clearly
illustrates the notion of `Functional Load as Information Loss'.
 The denominator is a normalizing factor
that makes it interpretable as the {\em fraction} of information lost
when the opposition is lost.

Other definitions of FL were also proposed by linguists, some
information theoretic e.g. \namecite{kucera} and some not
e.g. \namecite{greenberg}, \namecite{king}.

\subsection{Measuring constrasts' use in the Speech Recognition community}

Interest in FL among linguists waned after 1970. When it arose in a
different guise in the automatic speech recognition (ASR) community in
the 1980s, nobody noticed --- in either community.  Ironically, several
linguists had previously predicted that FL would be useful for ASR
research.

One reason that the connection was not spotted was due to the very
different way the concept originated in ASR. We now describe this. It
was thought possible to build broad-class recognizers for a language
$L$ that could tell with very high accuracy that a stop (or fricative
or vowel or...) had occurred, even if they could not recognize exactly
which stop it was. The hope was that this would be enough to recognize
most words. What was required was a measure of how well such a
recognizer worked, or at least an estimate of how well it would work
once it was made.

Such a recognizer could be represented by a partition $\theta$ of
phonemes whose classes were the broad classes it recognized
well. $\theta$ induces a partition $W_\theta$ of the set $W$ of words
in $L$. The elements of $W_\theta$ are word classes, or {\em cohorts}
in the notation used by \namecite{shipman}.  For example, if $\theta$
is the vowel-glide-other partition, the words `yak', `yap', `wit', etc
end up in one cohort, the words `chopping', `jotted', `fatten', etc in
another cohort, and so on. 

Several measures were proposed for the effectiveness\footnote
{
  The three proposed definitions summarized here all
  share the property that the higher they are, the worse the recognizer
  is. To be pedantic, they measure
  {\em in}effectiveness rather than  effectiveness. 
}
$e$ of a recognizer represented by a partition $\theta$. Since larger
cohorts are clearly worse, \namecite{shipman} proposed that
effectiveness be measured by the average cohort size: $e(\theta) =
\frac{1}{|W_\theta|}\sum_{C\in W_\theta} n(C)$, where $n(C)$ is the
number of words in cohort $C$. 
\namecite{huttenlocher} pointed out that this did not account for word
frequencies, and proposed that $e$ be the expected cohort size:
$e(\theta) = \sum_{C\in W_\theta} P(C)n(C)$. Note that $P(C)=\sum_{w\in
C} P(w)$ is the probability that a random word is in cohort $C$, where
$P(w)$ is the probability of word $w$. 
However, \namecite{carter} noted that this did not
adequately take into account word frequencies. He proposed that the
expected cohort entropy be used instead: $e(\theta) = \sum_{C\in
W_\theta} P(C)H(C)$. Note that the entropy $H(C) = - \sum_{w\in
C}\frac{p(w)}{p(C)}\log_2\frac{p(w)}{p(C)}$ of cohort $C$ is the uncertainty
in trying to tell apart words in it; it is harder to do so when  $H(C)$ is higher.

It turns out that Carter's definition of $e(\theta)$, the expected
uncertainty given that one can tell which cohort a word is in, is the
same as the conditional entropy given the same conditions,
i.e. $H(W|W_\theta) = H(W)-H(W_\theta)$. As this is not obvious, his
direct proof of it is reproduced below for completeness.

\begin{eqnarray*}
\sum_{C\in W_\theta} P(C)H(C) &=& - \sum_{C\in W_\theta} P(C)\sum_{w\in C}\frac{p(w)}{P(C)} \log \frac{p(w)}{P(C)} \\
&=& - \sum_{C\in W_\theta} \sum_{w\in C} p(w)\log p(w)   + \sum_{C\in W_\theta} \log P(C) \sum_{w\in C} p(w) \\
&=& - \sum_{w\in W} p(w) \log p(w) + \sum_{C\in W_\theta} \log P(C) \cdot P(C) \\
&=& H(W) - H(W_\theta)
\end{eqnarray*}


Carter's final measure was the Percentage of Information Extracted by $\theta$:

\begin{equation}\label{pie}
PIE(\theta) = \frac{H(W_{\theta}) }{ H(W) } 100\%
\end{equation}

$1-PIE(\theta) = \frac{H(W)-H(W_\theta)}{H(W)}$ looks very similar to
(\ref{FLhockett}); our framework includes both as special cases. It is
noteworthy that Carter does not cite Hockett's work, indicating that
he was not aware of it.


\section{Defining a framework}\label{FLdef}

We assume that a language is a sequence of discrete units, and that
the units can have a complicated structure. 

\subsection{Describing units}

A language $L$ is a sequence $L_{\tt T}$ of {\em objects} of
{\em type} {\tt T}, or {\tt T}-objects. For example, phonemes are objects
of type {\tt phn}. Each {\tt T}-object $x$ has a
{\em value} $v(x)$, which is one of a countable set $\Phi_{\tt T}$ of
possible values. For convenience, we shall often make references to
types implicitly, e.g. using $L$ for $L_{\tt T}$ and `object' instead
of `{\tt T}-object'. 


Types can be atomic or non-atomic. Non-atomic types are made using
atomic types and/or other non-atomic types. If {\tt T} is non-atomic,
then a {\tt T}-object $x$ is made of a positive number, say $n$, of components
$x_1,\ldots,x_n$, which are objects of type ${\tt T}_1,\ldots, {\tt T}_n$.
Its value $v(x)$ is the $n$-tuple $\langle v(x_1),\ldots,v(x_n)\rangle$ 
 of the values of its components, and must be one of
$\Pi_{j=1}^{n}\Phi_{{\tt T}_j}$. The set $\Phi_{\tt T}$ of all possible
values a {\tt T}-object can take is $\cup_{n=1}^\infty
\Pi_{j=1}^{n}\Phi_{{\tt T}_j}$.

Two {\tt T}-objects $x$ and $y$ are {\em equal} iff (if and only if)
they have the same value, i.e. $v(x)=v(y)$. If {\tt T} is atomic, it is
clear what this means. If {\tt T} is non-atomic, then $v(x)=v(y)$ iff
they have the same number of components and $v(x_i)=v(y_i)\ \forall
i=1,\ldots,n$. 

There are several ways in which non-atomic types can be formed; we
make use of only two.  In the first, and usual case, the number and
types of components in a {\tt T}-object depend only on its type. (Thus we
can associate components with types, rather than with objects.)
{\tt T}-objects all have the same number, $n({\tt T})$, of components, and
have one of the values in $\Phi_{\tt T} = \Pi_{j=1}^{n({\tt T})}\Phi_{{\tt T}_j}$.
The second case is for type {\tt string<T>}, where the number of
components can be any positive integer, but all components are of the
same type, {\tt T}.

For example, we could use the following system to represent a human
language as a sequence of words. A word is an object of type {\tt
wrd}, with two components, one of type {\tt syl} and another of type
{\tt mea}. {\tt mea} is an atomic type representing
`meaning'\footnote{ This paper never goes beyond phonology, so we
do not ever use such a type. }.
 A syllable is an object of non-atomic type {\tt syl}, and
has two components, of type {\tt string<phn>} and {\tt str}. {\tt phn}
is an atomic type representing phonemes, while {{\tt str}} is an atomic
type representing stress. If the language was tonal, syllables could
have a third component for tone.

More examples are given in Section \ref{egtypes}.

\subsection{Describing contrasts and their absence}\label{defncontrast}

It is not intuitively clear how to define a contrast in a
language. One reason for this is that contrasts are better described
by their absence than by their presence.  Suppose $c$ is some contrast
in language $L_{\tt T}$. There are several ways to define the process by
which $c$ is removed from $L_{\tt T}$; we choose one that works object by
object.

Consider the set $\Phi_{\tt T}$ of possible values of {\tt T}-objects. In the
absence of contrast $c$, some of the values will become
indistinguishable from other values. ``Equal in the absence of $c$''
is an equivalence relation that induces a partition, call it
$\theta_c$, on the set $\Phi_{\tt T}$ of possible values of
{\tt T}-objects. For example, suppose English is represented as a sequence
of phonemes (${\tt T}={\tt phn}$, $L_{\tt T}=English$) and $c$ is the voicing
contrast. Without voicing, phonemes like [t] and [d] sound identical,
as would [s] and [z], or [f] and [v], etc. This is represented by the
partition $\theta_{voicing}$ whose only equivalence classes with more
than one element are \{p,b\}, \{t,d\}, \{k,g\}, \{s,z\}, \{f,v\},
\{\xsh,\xzh\}, \{\xth,\xdh\} and \{\xch,\xdzh\}.

Just as $c$ defines $\theta_c$, so does any partition of $\Phi_{\tt T}$
define a contrast, i.e. $c\leftrightarrow\theta_c$.  We thus define a
contrast in a language $L_{\tt T}$ to be any partition of $\Phi_{\tt T}$.
Notationally, this means we can drop $c$ from our notation, and just
use $\theta$ to represent a contrast. $\theta$, being a partition of
$\Phi_{\tt T}$, is implicitly parametrized by ${\tt T}$. We will find it
useful to identify $\theta$ with the function
$g_{{\tt T},\theta}:\Phi_{\tt T}\rightarrow\theta$, where $g_{{\tt T},\theta}(v)$
is the equivalence class of $v$ in $\theta$.

Let us return to the question of what happens when a contrast $\theta$
disappears from $L_{\tt T}$. A new language $L_{{\tt T}_\theta}$ is created,
which is a sequence of ${{\tt T}_\theta}$-objects. ${\tt T}_\theta$ is a new
type that is defined to be just like {\tt T} in its component structure,
but its possible values are equivalence classes in $\theta$. Therefore:

\begin{equation}\label{TTtheta}
\Phi_{{\tt T}_\theta} = \theta
\end{equation}

As already mentioned, the function converting $L_{\tt T}$ to
$L_{{\tt T}_\theta}$ operates object by object. In other words, every
{\tt T}-object $x$ in $L_{\tt T}$ is replaced by a ${\tt T}_\theta$-object with
value $g_{{\tt T},\theta}(v(x))$.  Note that because of (\ref{TTtheta}),
$g_{{\tt T},\theta}$ is a function from $\Phi_{\tt T}$ to $\Phi_{{\tt T}_\theta}$ as
well.

Examples of contrasts are given in Section  \ref{egcontrasts}.

\subsection{The functional load of a contrast}

A language $L_{\tt T}$ is a sequence of {\tt T}-objects. If we assume that
$L_{\tt T}$ is generated by a stationary ergodic process, which we also
call $L_{\tt T}$, then its entropy $H(L_{\tt T})$ is well-defined, being the
entropy of its stationary distribution. The entropy of a distribution
$D$ over a countable set is $H(D) = - \sum_i p_i\log_2 p_i$, where
$p_i$ is the probability of the $i$-th member of $D$. Note that
$p_i\log_2 p_i$ is taken to be zero if $p_i=0$.

We define the functional load of a contrast $\theta$ in $L_{\tt T}$ as

\begin{equation}\label{fldefn}
FL_{\tt T} (\theta) = \frac{ H(L_{\tt T}) - H(L_{{\tt T}_\theta}) }{ H(L_{\tt T}) }
\end{equation}

In practice, we assume that the stationary ergodic
process is a very special process, namely a $(n-1)$-order Markov
process, which we denote by $L_{{\tt T},n}$. This means that the
probability distribution on the value of a {\tt T}-object depends on the
preceding $n-1$ {\tt T}-objects. The entropy of $L_{{\tt T},n}$, 
which is the entropy of the distribution of $n$-grams of {\tt T}-objects, 
is an $n$-th order approximation to that of $L_{\tt T}$ that improves as
$n$ becomes larger; \namecite{shannon} proved that $H(L_{\tt T}) =
\lim_{n\rightarrow\infty} H(L_{{\tt T},n})$. 

We may want to bear in mind a passing
comment by \namecite{hockett67}.  He suggested that finite $n$ might
actually be more appropriate for languages, as articulatory
constraints prevent the formation of infinitely long
utterances. Perceptual mechanisms clump phonemes into cohesive units,
such as syllables or words, when presented with long utterances. In
principle, clumping never stops; sequences of words get clumped into
sentences, and so on. How far the assumption of generation by a
stationary, ergodic Markov process can be taken is not known. 

We define the $n$-th order approximation to the functional load of 
contrast $\theta$ in $L_{\tt T}$ as

\begin{equation}\label{fldefn_n}
FL_{{\tt T},n} (\theta) = \frac{ H(L_{{\tt T},n}) - H(L_{{\tt T}_\theta},n) }{ H(L_{{\tt T},n}) }
\end{equation}

Note that taking ${\tt T}={\tt phn}$ gives Hockett's 
formula (\ref{FLhockett}) while taking ${\tt T}={\tt wrd}$, with $n=1$
fixed, gives Carter's formula (\ref{pie}).

The parameters of $L_{{\tt T},n}$ must be estimated using a finite sample
of its outputs, i.e. a finite sequence of {\tt T}-objects. This finite
sequence is called a {\em corpus}. We denote by $\HH(L_{{\tt T},n};S)$ the
entropy of the process $L_{{\tt T},n}$ when its parameters are estimated
using corpus $S$. $N$, the number of {\tt T}-objects in $S$, and the
structure of $\Phi_{\tt T}$, determine how large $n$ can be made
before sparse sampling problems become an issue.

There are several ways of finding the estimate $\HH(L_{{\tt T},n};S)$ from
$S$. We used the classical method of normalized counts of $n$-grams in
$S$.  Suppose $c(u_1\ldots u_n)$ is the number of times $u_1\ldots
u_n$ (each $u_i\in\Phi$) appears as a contiguous subsequence of
$S$. Define a probability distribution $D_{n}$ over $n$-grams by
$p(u_1\ldots u_n)=\frac{c(u_1\ldots u_n)}{N-n+1}$.  Then $\HH(L_{{\tt
T},n};S) := \frac{1}{n}H(D_{n})$.

To illustrate, consider a toy language $L$ represented by a sequence of
{\tt toy}-objects with $\Phi_{\tt toy}=\{$a,b,c\}. The corpus to be
used is $S = $`abaccaaccaabbacabab'. Say $n=2$.  The distribution
$D_2$ of {\tt toy} bigrams in $S$ is (aa 2), (ab 4), (ac 3), (ba 3),
(bb 1), (bc 0), (ca 3), (cb 0), (cc 2). $H(D_2) = - \frac{2}{18}
\log_2 \frac{2}{18} - \frac{4}{18} \log_2 \frac{4}{18} - \ldots -
\frac{2}{18} \log_2 \frac{2}{18} = 2.7108.$ So $\HH(L_{{\tt toy},2};S)
= {\frac12}2.7108=1.3554$.

This means that our estimate of the $n$-th order approximation to the
functional load of a contrast $\theta$ in $L_{\tt T}$ is 

\begin{equation}\label{fldefn_nS}
\FL_{{\tt T},n} (\theta; S) = \frac{ \HH(L_{{\tt T},n};S) - \HH(L_{{\tt T}_\theta,n}; g_{{\tt T},\theta}(S)) }{ \HH(L_{{\tt T},n}; S) }
\end{equation}

For convenience, we will often write $FL_{{\tt T},n,S}(\theta)$ for $\FL_{{\tt T},n}(\theta;S)$.
\\

Let us return to the toy example. If we do
not make use of the b/c opposition, any occurrence of b or c in
the corpus $S=$`abaccaaccaabbacabab' is taken to be an occurrence of
the same symbol, which we call, say, d. The corresponding partition
$\theta_{bc}$ of $\Phi_{\tt toy}$ is $\{\{a\},\{b,c\}\} \simeq \{a,d\}
= \Phi_{\theta_{bc}}$. The converted corpus $g_{{\tt toy},\theta_{bc}}(S)$ reads
`adaddaaddaaddadadad'. The distribution of ${\tt toy}_{\theta_{bc}}$
bigrams is (aa 2), (ad 7), (da 6), (dd 3) and the resulting entropy
$1.8016$. Plugging these values in (\ref{fldefn_nS}) gives $\FL_{{\tt
toy},2} (\theta_{bc};S) = \frac{2.7108 - 1.8016}{2.7108} = 0.335$,
meaning that the b/c contrast carries over a third of the information
in $S$ --- when $n$ is 2.

Clearly, there are two nuisance parameters here, $n$ and $S$. In
section \ref{testconsistency}, we investigate how much difference the
choices of $n$ and $S$ makes. We find they do not make as much
difference as might be feared, possibly since the entropies in the
numerator and denominator 'cancel out'. However, they are still
certainly an issue to keep in mind, and a few remarks on them are in
order.

Most linguists, when speaking of phonological rules, usually assume
$n=1$, going to $n=2$ for a few rules involving word boundaries. This
is both because many rules don't go beyond two word boundaries and
because it is convenient to do so. In other words, the approximations
we make here are no worse than those usually made by linguists.

That the choice of $S$ makes a difference is clear; the entropy of a
text can even be used to distinguish between authors
\cite{kontoyannis} writing in the same language. We suspect, without
proof, that FL is more robust than entropy to changes in $S$, since FL
normalizes entropy both additively and multiplicatively.


\section{What types to use for human languages}\label{egtypes}

\subsection{Non-tonal languages}

In the calculations for Dutch, English, and German in Section
\ref{difflang}, we used four types, for phonemes, stress, syllables
and words. The first two types are atomic. All $\Phi_{\tt T}$ differ
with language; the examples given here are for English.


\begin{itemize}

\item Objects of type {\tt phn}, which we call phonemes for
convenience, take values in $\Phi_{\tt phn} =
\{$[p],[t],[k],$\ldots$,[{\xae}],[i],[{\xI}]$\}$. 

\item {\tt str}-objects take values in
$\Phi_{\tt str}=\{$primary, secondary, unstressed$\}$. 

\item {\tt syl}-objects (syllables) have two components; $n(${\tt
syl}$)=2$. The first is of type {\tt string<phn>} and the second of
type {\tt str}. Two syllables with values 
$\langle$mi{\xng},unstressed$\rangle$ and
$\langle$mi{\xng},primary$\rangle$ are not equal, since although their phonemic
 components are equal, their stress components are not.

\item {\tt wrd}-objects (words) have a single component, of type {\tt string<syl>}. 

\end{itemize}

\subsection{Tonal languages}

In the calculations for Mandarin and Cantonese in this chapter, we
used the same setup as for the non-tonal languages, bar two
changes. First, of course, the sets of possible values ($\Phi_{\tt
phn}$, $\Phi_{\tt wrd}$, etc) differ with language.  Second, syllables
have an additional component for tone, of atomic type {\tt ton}. In
Mandarin, for example, the set of possible tonal values is $\Phi_{\tt
ton}= \{$high level, rising, low level, falling, no tone$\}$.

Of course, allocating tones to syllables is an idealization, since
tone sandhi and coarticulation occur in continuous speech.  An example
of the former, due to \namecite{chao}, is with the words `yi', `qi',
`ba' and `bu' which have high, high, high and falling tone in
isolation\footnote{These words are written in Pinyin. They mean `one',
`seven', `eight' and `no' respectively in English.}. In continuous speech
they all have falling tone unless they are followed by a falling tone,
in which case they have a rising tone. Such cases are predictable in
that they could be corrected for with corpus pre-processing. However,
we did not correct for them.

Regarding coarticulation, \namecite{xudiss} found that ``Mandarin speakers
identify the tones presented in the original tonal contexts with high
accuracy. Without the original context, however, correct
identification drops below chance for tones that deviate much from the
ideal contours due to coarticulation. When the original tonal context
is altered, listeners compensate for the altered contexts as if they
had been there originally. These results are interpreted as
demonstrating listeners' ability to compensate for tonal
coarticulation.'' While this justifies our idealization to a large
extent, bear in mind that the compensation for coarticulation is by no
means perfect, particularly where adjacent tones `disagree'
\cite{xudiss,xucoartictone}.

\subsection{Extensions required}

The model of phonology used in this paper is more general than
classical structural phonology. However, one may well ask how we could
make use of more sophisticated models such as autosegmental phonology
\cite{goldsmith}, especially since a computational framework for it
already exists \cite{amar}. 

We are not sure how this can be done. However, we have some
suggestions, which involve making components correspond to tiers.  We
need to assume that there is some overall (i.e. over all tiers) unit
that no object in any tier ever straddles. For example, in a language
where a tone can be associated with vowels in different words, such a
unit would have to be strictly larger than a word. Even so, taking it
to be a word still permits several phonological rules to be
represented as contrasts (see Section \ref{phonrules}). Among the
details we have yet to sort out is how to represent association lines
between tiers.

\begin{figure}
\centerline{\epsfig{file=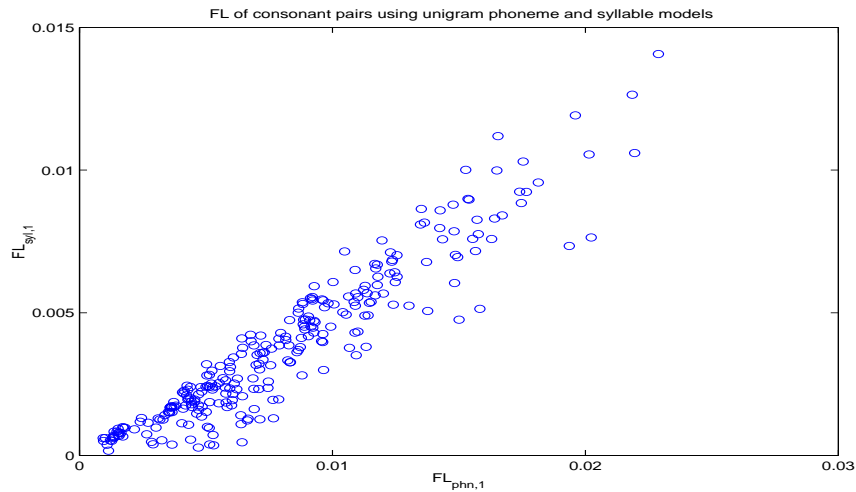,width=13cm,height=7cm}}
\caption{Comparing the  FL of 276 consonant pairs using phoneme unigrams and
syllable unigrams, in the Switchboard corpus. The correlation is 0.942.}
\label{phn1_syl1_conspairs}
\end{figure}


\section{Examples of contrasts}\label{egcontrasts}

In Section \ref{defncontrast}, any partition of $\Phi_{\tt T}$
defines a contrast in a language represented as a sequence of
{\tt T}-objects. This allows us to use the word `contrast' in a more
general sense than is standard, as the examples in this section
show. These examples make use of the types defined in Section \ref{egtypes}.

\subsection{Phoneme oppositions}

Nearly all previous work on FL, in both linguistics and speech
recognition, has been on phoneme oppositions, especially binary.

Suppose a language is a sequence of phonemes. Almost any phoneme
opposition can be represented by a partition $\theta$ of $\Phi_{\tt
phn}$ with the opposition being that between phonemes in the same
equivalence class of $\theta$.  For example, the binary opposition of
phonemes $x$ and $y$ is represented by $\theta$ being the partition
with just one non-singleton equivalence class, $\{x,y\}$.

More generally, if the opposition is between phonemes in set
$A\subseteq\Phi_{\tt phn}$ then we can take $\theta$ to be the
partition of $\Phi_{\tt phn}$ with $A$ as one
equivalence class and all other classes with one phoneme
each. $A=\{x,y\}$ is, of course, the binary opposition case of the
previous paragraph.  Note that the contrast here is `distinguishing
between phonemes within $A$', not `distinguishing phonemes in
$A$ from phonemes not in $A$'. Table \ref{FL_groups_difflangs} has some
examples. 

Even more generally, if the opposition is between phonemes in several
pairwise-disjoint sets of phonemes, take $\theta$ to be the partition
defined by these sets. For example, if the opposition is between
consonants and between vowels simultaneously, take $\theta$ to be the
two-class partition of consonants and vowels. $FL(\theta)$ then
represents the information lost when one can tell whether a consonant
or vowel has occurred, though not which vowel or which consonant. 

This is all very well if {\tt T} is in fact {\tt phn}. But what if the
objects are syllables or words? In this case, we make use of
inheritance across types. For example, if ${\tt T}={\tt syl}$,
since syllables have a {\tt string<phn>} component, any partition of
$\Phi_{\tt phn}$ induces a partition of $\Phi_{\tt syl}$.
Similarly, if ${\tt T}={\tt wrd}$, since words have a {\tt string<syl>}
component, any partition of $\Phi_{\tt phn}$ induces a partition of
$\Phi_{\tt syl}$ which in turn induces one of $\Phi_{\tt
wrd}$. Thus partitions of $\Phi_{\tt phn}$ are contrasts whether the
objects are phonemes, syllables or words. 

This is better explained if we use $g_{{\tt T},\theta}$ instead of
$\theta$. Recall from Section \ref{FLdef} that $g_{{\tt T},\theta}$ is the
function converting the original language $L_{\tt T}$ to the contrast-less
language $L_{{\tt T}_\theta}$ by sending all {\tt T}-objects with values in
the same equivalence class of $\theta$ to a ${\tt T}_\theta$-object with
the same value. For example, suppose, once again, that the contrast is
between phonemes in some set $A$ and that ${\tt T}={\tt phn}$. For any
phoneme $p\in\Phi_{\tt phn}$,

\begin{equation}\label{phnA}
g_{{\tt phn},\theta}(p) = \left\{
\begin{array}{ccl}
A & {\rm if } & p\in A \\
p & {\rm if } & p\not\in A 
\end{array}
\right.
\end{equation}

For convenience, we abuse notation by mapping $p$ to itself, rather
than to $\{p\}$, if $p\not\in A$. 

Now, suppose ${\tt T}={\tt syl}$ and 
$\theta$ is the same partition of $\Phi_{\tt phn}$. Syllables have a
component of type {\tt string<phn>}; for concreteness, suppose they have only one other component, of type {\tt str}. 
Thus, a typical syllable is an ordered 2-tuple
$\langle p_1\ldots p_m,s\rangle$, where each $p_i\in\Phi_{\tt
phn}$ and $s\in\Phi_{\tt str}$. Now we have

\begin{equation}\label{sylA}
g_{{\tt syl},\theta}(\langle p_1\ldots p_m,s\rangle) = 
\langle g_{{\tt phn},\theta}(p_1)\ldots g_{{\tt phn},\theta}(p_m),s\rangle
\end{equation}

Notice that until now, $\theta$ had to be a partition of
$\Phi_{\tt T}$. However, now ${\tt T}={\tt syl}$, but $\theta$ is a partition of
$\Phi_{\tt phn}$. This is not a contradiction, but merely systematic abuse
of notation, since any partition of $\Phi_{\tt phn}$ naturally induces a
partition of $\Phi_{\tt syl}$. 

If $\theta'$ is another partition of $\Phi_{\tt str}$,
represented by a function $h_{{\tt str},\theta'}$, then $\theta$ and
$\theta'$ applied simultaneously result in a contrast represented by
a function taking $\langle p_1\ldots p_m,s\rangle$ to $\langle g_{{\tt
phn},\theta}(p_1)\ldots g_{{\tt phn},\theta}(p_m),h_{{\tt
str},\theta'}(s)\rangle$.

\subsection{Distinctive Features}\label{distfeat}

By distinctive feature, we refer to characteristics used to
distinguish phonemes, such as aspiration, voicing, place, manner,
etc. Distinctive features do not have to be binary.

Any distinctive feature can be represented by a partition $\theta$ of
$\Phi_{\tt phn}$ which has two or more phonemes in the same class
iff they would be merged in the absence of the feature. For example,
if voicing were lost in English, $\theta$ is  $\theta_{voicing}$ in
Section \ref{defncontrast},  where  [t] and [d] are in one
equivalence class, [s] and [z] in another, [{\xsh}] and [{\xzh}] in
another, etc, with all other phonemes in their own classes. 

Most well-studied languages have several possible organizations of its
phonemes and distinctive features\footnote{The number of organizations
is a monotonically increasing function of the number of studies of the
language. The nature of this function requires, though not necessarily
deserves, further study.} Any organization can
be used, as long as one is specified. 
What we
mean by organization is best explained by example; we used the
organizations in Tables \ref{mandarin_featvals} and \ref{DEG_featvals}
for Mandarin, Dutch, English and German to get the FL of different
features in each language in Table \ref{FL_features_difflangs}.

\subsection{Suprasegmental contrasts}

Suppose we model a language by a sequence of syllables, with each
syllable having a stress component. Since any partition of
$\Phi_{\tt str}$ induces one of $\Phi_{\tt syl}$ by inheritance,
any partition $\theta$ of $\Phi_{\tt str}$ is a contrast. This
remains the case if we model a language by a sequence of words where
words have a {\tt string<syl>} component, since any partition of
$\Phi_{\tt syl}$ induces one of $\Phi_{\tt wrd}$. 

To find the FL of stress, use the partition of $\Phi_{\tt str}$ with
a single class containing all stress values. This is equivalent to not
having any information about stress at all.

Suppose we were dealing with a language like English with different
kinds of stress, and we wanted to find out how importance it was to be
able to distinguish primary from secondary stress. Then we would use
the partition \{\{primary,secondary\},\{absent\}\} of $\Phi_{\tt
str}$. If we wanted to find out how importance it was to distinguish
secondary stress from no stress at all, we would use
\{\{primary\},\{secondary,absent\}\} instead.

If we were modelling a tonal language, with syllables having a tonal
component, then everything above said for stress would apply to
tone, with tonal contrasts represented by partitions of $\Phi_{\tt
ton}$. For instance, to find the FL of tone, use the 1-class
partition of $\Phi_{\tt ton}$.

\subsection{Phonological rules}\label{phonrules}

In all the previously described contrasts, the conversion from {\tt
T}-object to {\tt T}-object was absolute, i.e. it happened in every
situation where it could happen. Sometimes, we would like the
conversion to occur only in certain situations. 

For example, if we wanted to find the functional load of vowels when
${\tt T}={\tt syl}$, we would take $\theta$ to be the partition of
$\Phi_{\tt phn}$ whose only non-singleton equivalence class was $V$,
the set of vowels. Defining $g_{{\tt phn},\theta}$ as in (\ref{phnA}), we would write, as in (\ref{sylA})

$$
g_{{\tt syl},\theta}(\langle p_1\ldots p_m,s\rangle) = 
\langle g_{{\tt phn},\theta}(p_1)\ldots g_{{\tt phn},\theta}(p_m),s\rangle
$$

Now, suppose we wanted to represent the contrast of vowel reduction,
i.e. of not being able to distinguish between vowels in unstressed
syllables.  This means that every vowel is replaced by a single vowel
placeholder, but only if the syllable containing it is unstressed. In
other words, the mapping is now:

$$
g_{{\tt syl},\theta}(\langle p_1\ldots p_m,s\rangle) = 
\left\{
\begin{array}{ccl}
\langle g_{{\tt phn},\theta}(p_1)\ldots g_{{\tt
phn},\theta}(p_m),s\rangle  & {\rm if } & { s\ }{\rm is\ unstressed} \\
\langle p_1\ldots p_m,s\rangle & {\rm if } & {\rm not} 
\end{array}
\right.
$$

where

\begin{equation}\label{phnV}
g_{{\tt phn},\theta}(p) = \left\{
\begin{array}{ccl}
V & {\rm if } & p {\rm \ is\ a \ vowel }\\
p & {\rm if } & p {\rm \ is\ not \ a \ vowel }\\
\end{array}
\right.
\end{equation}

Some phonological rules in linguistics fit in this framework very
nicely. For example, epenthesis of [t] in the consonant cluster [n\_s]
in English is represented by the function

$$
g_{{\tt syl},\theta}(\langle p_1\ldots p_m,s\rangle) = 
\left\{
\begin{array}{ccl}
\langle p_1\ldots p_i {\rm [t]} p_{i+1}\ldots p_m,s\rangle & {\rm if } &
p_i={\rm [n]} \ \& \ p_{i+1}={\rm [s]} \\
\langle p_1\ldots p_m,s\rangle & {\rm if } & {\rm not} 
\end{array}
\right.
$$

In this case, $\theta$ corresponds to the partition of $\Phi_{\tt
syl}$ where two syllables are in the same equivalence class iff
$g_{{\tt syl},\theta}$ maps them to the same value. Thus the syllables
[k{\xae}nts] and [k{\xae}ns] end up in one class, [l{\xI}ns] and
[l{\xI}nts] in another, and so on. If ${\tt T}={\tt wrd}$ instead,
then words like `tense' and 'tents' would end up in the same class,
`mince' and `mints'  in another, and so on.

\subsection{The contrast of a single phoneme}\label{FLsinglephoneme}

At first, it makes little sense to speak of the functional load of a
single phoneme. After all, phonemic oppositions require at least two
phonemes to be in opposition.

A clue to how to proceed is given by \namecite{ingram}, who states
that the FL of [\xdh] in English must be low because ``we could change
all English /dh/ into [d]'s and still communicate''. He was referring
to the fact that /dh/, which is the most frequent consonant in
English, does not intuitively seem to be most relied-upon consonant. 

More generally, the question to be asked is `how can a phoneme
disappear from a language?' Some phonemes, like [h] in Cockney
English, disappear. Others vanish by merging with other phonemes,
e.g. [n] with [l] in Cantonese. The merger need not be absolute,
i.e. with the same phoneme everywhere, of course.

We define the contrast of a single phoneme to be the phonological rule by
which the phoneme disappears from the language. Therefore $FL(x)$ is
the FL of the phonological rule for the disappearance of phoneme $x$.

Unfortunately, the process by which a phoneme disappears can rarely be
predicted before it, if it ever does, disappears.  What is needed is a
comprehensive survey of how a given phoneme has disappeared from
various languages in the past. Such a survey would be able to answer
hypotheses like 'does /h/ ever disappear by a process other than
deletion?', or `do phonemes only merge with phonemes that share the
same place (phonemes with secondary articulations being considered as
having two places of articulation)?'

Our current working definition for $FL(x)$, in the case of disappearance-by-merger, is as follows.  Suppose $x$
can only potentially merge with phonemes in a set $S(x)$ of phonemes
`similar' to it, and that the probability that it merges with phoneme
$y\in S(x)$ is $P(x,y)$. Then 

$$FL(x) =  \sum_{y\in S(x)-x} P(x,y) FL(x,y) $$

\begin{figure}
\centerline{\epsfig{file=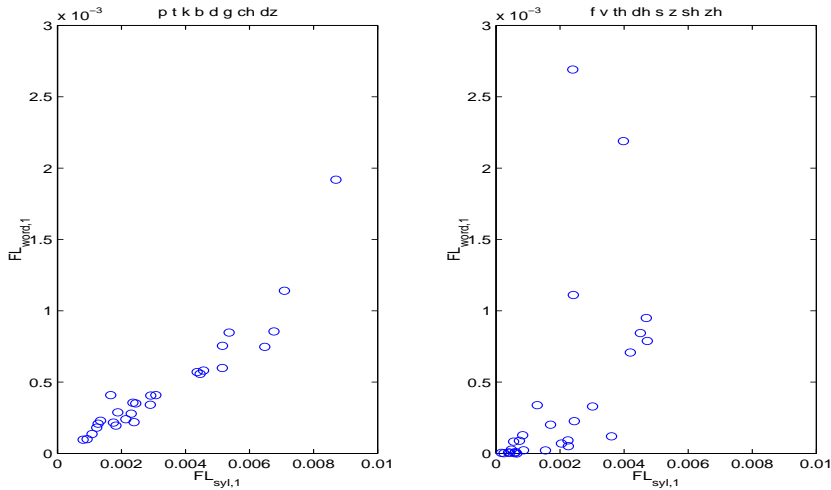,width=13cm,
       height=7cm}} 
\caption{Comparing Functional Load values for 28 pairs of obstruent consonants
using unigram syllable (horizontal axis) and word  based
computations. Both are based on the CELEX lexicon. The left plot is
for pairs from \{p,t,k,b,d,g,{\xch},{\xdzh}\}; the correlation is 0.927.  The
right plot is for pairs from \{f,v,{\xth},{\xdh},s,z,{\xsh},{\xzh}\}; the 
correlation is 0.611.  Both plots are to the same scale; the
horizontal axis is from 0 to 0.010 while the vertical is from 0 to
0.003.}
\label{FL_stopsfrics_sylword_celex}
\end{figure} 

This can be interpreted as the expected FL of $x$, taken over possible
absolute mergers. Alternatively, it can be interpreted as the FL of the process where $x$ merges with
phonemes in $S(x)$, merging with different phonemes in different
environments such that $P(x,y)$ is the proportion of environments
where $x$ merges with $y$.


\section{The robustness of the measure}\label{testconsistency}

\subsection{Measuring robustness}

We would like to speak of $FL_{\tt T}(\theta)$ without reference to
the parameters $n$ and $S$. This cannot be done if we expect any two
possible measures to give the same {\em absolute} value for any
contrast. For example, for most contrasts $\theta$,
$\FL_{{\tt T},n}(\theta; S)$ will be larger than $\FL_{{\tt T},n+1}(\theta;
S)$ because larger $n$-grams capture more information.  Instead we
wish them to give the same `relative' values, to be highly predictable
from each other.

We define measures $FL_1$ and $FL_2$ to be {\em consistent} for a
set ${\bf\Theta}$ of contrasts iff there is a constant
$\gamma_{12}$ such that $FL_1(\theta) = \gamma_{12} FL_2(\theta) \
\forall \theta\in{\bf\Theta}$.

In practice, we can only hope for $FL_1(\theta) \approx \gamma_{12}
FL_2(\theta)$. Bearing in mind that what is important is not the value of $\gamma_{12}$ but its existence, we define
 $\alpha_{\bf\Theta}(FL_1,FL_2)$ to be the linear (Pearson's) correlation between values $FL_1(\theta)$ and $FL_2(\theta)$,
when $\theta$ is taken over all values in ${\bf\Theta}$. In other
words, 

$$\alpha_{\bf\Theta}(FL_1,FL_2) = 
\frac{1}{|{\bf\Theta}|}
\sum_{\theta\in{\bf\Theta}}
Z(FL_1(\theta)) Z(FL_2(\theta)) 
$$

Note that $Z(FL_i(\theta)) = \frac{FL_i(\theta)-\mu_i}{\phi_i}$, where $\mu_i
= \frac{1}{|{\bf\Theta}|}\sum_{\theta\in{\bf\Theta}}
FL_i(\theta)$ and \\
$\phi_i^2 = \frac{1}{|{\bf\Theta}|-1}\sum_{\theta\in{\bf\Theta}}
(FL_i(\theta)-\mu_i)^2$.

The maximum, ideal, value of $\alpha_{\bf\Theta}$ is 1.  We do not
know how high it must be for $FL_1$ and $FL_2$ to be consistent in
general, though we have rules of thumb for specific cases.

This section gives evidence for the
consistency of $\FL_{{\tt T},n}(\theta;S)$ and $\FL_{{\tt
T},n'}(\theta;S')$ for different $n,n'>0$ and corpora $S,S'$.  This restriction in the
interpretation of FL still allows it to be useful, as described in
Section \ref{interpret}.

\subsection{Testing Procedure}

Unless otherwise specified, we will restrict ourselves to a limited
collection of contrasts, namely binary oppositions. These are very fine
contrasts (i.e. the partition of $\Phi_{\tt phn}$ they rely on is
almost the finest possible) and consistency for them is indicative of
consistency for other contrasts.

Suppose that $\Phi_0 \subset \Phi_{\tt phn}$ is a subset of phonemes, and
${\bf\Theta_{\Phi_0}}$ is the set of all contrasts that are binary
oppositions of pairs of phonemes in $\Phi_0$. For example, if
$\Phi_0=\{w,x,y,z\}$, then ${\bf\Theta_{\Phi_0}}$  is $\{\theta_{wx},
\theta_{wy}, \theta_{wz}, \theta_{xy}, \theta_{xz}, \theta_{yz} \}$. 
 For convenience we
define $\alpha_{\Phi_0}(FL_1,FL_2)$ to be $\alpha_{\bf
\Theta_{\Phi_0}}(FL_1,FL_2)$.  All correlations reported here are
extremely significant, having $p<10^{-5}$ unless reported otherwise.
Our rule of thumb is that $FL_1$ and $FL_2$ are
consistent over $\Phi_0$ if $\alpha_{\Phi_0}(FL_1,FL_2)>0.9$.

While our testing was only done with English corpora, results should
hold for other languages. The corpora used were CELEX \cite{celex} and
Switchboard \cite{swb}. CELEX is essentially a word-frequency list
with each word having a citation form pronunciation and the frequency
with which it appears in the 16 million word (24 million syllables)
Birmingham/COBUILD corpus of British English.  Switchboard (SWB) is a
large 240-hour speech corpus, but we used the small but
extra-carefully transcribed ISIP subset of it\footnote{ Our thanks to
the researchers at Mississippi State who have made this subset freely
available at www.isip.msstate.edu/projects/switchboard}, which has 80
000 phonemes in 36 500 syllables in 2 hours of spontaneous telephone
speech by American English speakers.

\subsection{Consistency for different $n$}\label{consistency_n}

For any fixed {\tt T}, corpus $S$, $\Phi_0\subseteq \Phi$, we want
$\alpha_{\Phi_0}(FL_{{\tt T},m,S},FL_{{\tt T},n,S})$ to be as close
to 1 as possible for any positive integers $m,n$. Table
\ref{swb_FLcorr_difmarkov} shows its value when {\tt T} $=$ {\tt phn},
$S$ is Switchboard, $\Phi_0$ consists of all consonants (values for
vowels are higher) and $1\leq m,n\leq 5$. The correlation decreases
with $|m-n|$ but remains high throughout.

\begin{table}[h]
\begin{center}
\begin{tabular}{|c|llll|}
\hline
	& 1	& 2	& 3 	& 4\\
\hline
2	& 0.985 & 	& 	&\\
3	& 0.956 & 0.988 & 	&\\
4	& 0.928 & 0.961 & 0.988 &\\
5	& 0.878 & 0.906 & 0.947 & 0.978 \\
\hline
\end{tabular}
\end{center}
\caption{The correlation $\alpha_{consonants}$ between
$FL_{phn,n,Switchboard}$ for different $n$. 
}
\label{swb_FLcorr_difmarkov}
\end{table}

Similar results are found when ${\tt T}={\tt syl}$; 
$\alpha_{consonants}(FL_{{\tt syl},1,SWB},FL_{{\tt syl},2,SWB}) = 0.945$. However, 
sparsity concerns about the small size of Switchboard made values of
$FL_{{\tt syl},n,SWB}$ for $n>2$ suspect and larger values
of $n$ were not tried.

For {\tt T} = {\tt wrd}, we used frequency and sequence information from the
Brown corpus and pronunciation information from CELEX. We then
computed $FL_{{\tt wrd},n,Brown-CELEX}$ values for 200 randomly generated
partitions of $\Phi_{\tt phn}$, for $n=1,2,3$, and found that the correlation
was over 0.95 in each case.

We conclude from this that taking $n=1$, i.e. estimating FL with
unigrams, is adequate for many purposes. In the rest of this paper,
$n$ is 1 if not specified.

\subsection{Consistency for different corpora}\label{consistency_corpora}

For any fixed type {\tt T}, $n>0$, and $\Phi_0\subseteq \Phi_{\tt
phn}$, we want $\alpha_{\Phi_0}(FL_{{\tt T},n,S},FL_{{\tt T},n,S'})$
to be as close to 1 as possible for different corpora $S,S'$.  Taking
advantage of the results of Section \ref{consistency_n}, we assume
$n=1$.

We deal with {\tt syl} objects. The corpora in question are
Switchboard and CELEX. Note that stress information was removed from
CELEX for this comparison, since Switchboard syllables do not have
stress information\footnote{ Gina Levow informed us that syllables are
marked with stress in another subset of Switchboard. However, this was
after the calculations in this paper were done. }.  $\alpha_{consonants}$ $(FL_{{\tt syl},SWB},$
$FL_{{\tt syl},CELEX})$ $ = 0.826$ while $\alpha_{vowels}$ $(FL_{{\tt
syl},SWB},$ $FL_{{\tt syl},CELEX})$ $ = 0.730$.
Interestingly,
some consonants fare better than others: 
$\alpha_{obstruents}$ $(FL_{{\tt syl},SWB},$ $FL_{{\tt syl},CELEX}) = 0.920$
while 
$\alpha_{non-obstr.~cons.ts}$ $(FL_{{\tt syl},SWB},$ $FL_{{\tt syl},CELEX})$ is
0.762.
More details of this experiment are in Section \ref{nonideal}.

Although entropy is known to be very corpus dependent, it appears that
the normalized differences in entropy are more well-behaved. This is
certainly the case when obstruents are involved, in which case FL
calculations are robust. Other contrasts require further work, though
the computation of their FL is robust enough for many purposes.

\subsection{Consistency for different objects}

Object type is a necessary parameter when computing FL. Intuitively,
we expect some consistency for different types, but not in the same
way as for $n$ and $S$, and therefore inconsistency across different
types indicates interesting word structure patterns. In other words,
comparisons of $FL_{\tt T,n,S}$ and $FL_{\tt T',n,S}$, for different
{\tt T} and {\tt T}$'$,  could prove to
be a useful tool for linguistic analysis. 

We compare {\tt phn} and {\tt syl}, for $n=1$ and $S=$SWB. In this
case, $\alpha_{consonants}$ $(FL_{\tt phn},$ $FL_{\tt syl}) = 0.942$,
which is very high. The corresponding values for $\alpha_{vowels}$ is
even higher.  The surprise here is that $FL_{\tt phn}$ is based on
phoneme unigrams, i.e. how many times each phoneme appears, and thus
makes no use of context.

We compare {\tt syl} and {\tt wrd} with $n=1$ and $S=$CELEX. Here,
context turns out to be more important; $\alpha_{vowels}$ $(FL_{\tt syl},$ $FL_{\tt wrd})$ is 0.752
and $\alpha_{obstruents}$ $(FL_{\tt syl},$ $FL_{\tt wrd})$ is 0.716.
Interestingly, the latter figure really has two parts
(see Figure \ref{FL_stopsfrics_sylword_celex}) since
$\alpha_{stops+affricates}$ is 0.927
while $\alpha_{fricatives}$ is 0.611 ($p=0.001$). 
We do not know why this is so, nor why the latter figure (again) has two
parts, with $\alpha$ higher for voiced fricatives than unvoiced. 



\section{Computing FL with non-ideal data}\label{nonideal}

Robust FL computation means we can find usable FL values for languages
for which inadequate data is available. For example, there are
relatively few corpora that are manual phonetic transcriptions of `the
language as spoken'; this is particularly true for languages for which
there are few or no native speakers. On the other hand, word-frequency
pairs, with citation form pronunciations of words and frequencies
based on {\em written texts}, are easier to find.  To see if we can
accurately estimate FL using word-frequency pairs, we look at the
CELEX vs Switchboard calculations of Section \ref{testconsistency} in
more detail. These corpora represent opposite ends of several
spectrums, which makes for a good test.  The differences between them
are summarized here:

\begin{itemize}
\item Switchboard and CELEX reflect different dialects, American and
British respectively, of English.

\item The frequencies in CELEX are mostly based on written sources.

\item As CELEX gives word-frequency lists, all syllabifications in it
are word-internal or at word boundaries, unlike Switchboard.

\item CELEX reflects a much ($>$600 times) larger  corpus than Switchboard.

\item CELEX gives citation form pronunciations for each word. 30\% of words also have other pronunciations, but there is (unsurprisingly) little
information on how often each other pronunciation is used. The
word-frequency list we extracted from CELEX assigned a single
pronunciation to a word. This was the citation form except when
other pronunciations were available, in which case we took the most
common colloquial form.

\item 
Each syllable in CELEX is marked as having one of three types of
stress: primary, secondary and none. The syllable in monosyllabic
words has primary stress. Syllables in our Switchboard data are not
marked with stress. To make syllables
comparable, the stress component was removed from the CELEX syllables. 

\end{itemize}

At first sight, it would seem that we should compare $FL_{{\tt
wrd},CELEX}$ with $FL_{{\tt wrd},SWB}$. But this requires making the
sorts of assumptions (syllables don't cross word boundaries, same
pronunciation each time) about Switchboard as for CELEX, the very
assumptions we wish to test. To get an idea of what words look like in
continuous speech, consider the ARPABET-transcribed SWB sentence
below. Syllables are within square brackets and interphoneme silences
have been removed.

\texttt{ [l ay] [k ih n] [ao] [g ix] [s w eh] [n eh r] [iy] [b aa] [d
iy] }\\ {\tt [z aa n] [v ey] [k ey] [sh ih] [n er] [s ah m] [th ih ng
k] [w iy] }\\ {\tt [k ix n] [d r eh] [s el] [l el] [m ao r] [k ae] [zh
w ax l] }

The actual sentence is {\it ``Like in August when everybody is
on vacation or something we can dress a little more casual''}. 
Notice how often syllables cross word boundaries. 

Even if we weaken the restriction so that words are pronounced in 
a limited set of ways, it is hard to draw the line on what `limited'
means. Therefore, we shall instead compare $FL_{{\tt syl},CELEX}$ with
$FL_{{\tt syl},SWB}$. Then $\alpha_{\Phi_0}(FL_{{\tt syl},SWB},FL_{{\tt syl},CELEX})$ is
0.730, 0.826 and 0.920 for vowels, consonants, and
obstruents respectively.

We conclude that non-ideal corpora can give results consistent with
ideal corpora that are very representative of speech for contrasts
that involve consonants, particularly obstruent consonants.


\section{An application in linguistic typology}\label{difflang}

\begin{table}[h]
\begin{center}
\begin{tabular}{|l|c|c|c|c|c|c|}
\hline
		& Labial	& Alveolar	& Alv-pal 	& Retroflex	& Lateral & Velar \\
\hline
Stop		& p \{p$^{\rm h}$\} [m]	& t \{t$^{\rm h}$\} [n]	&		&		&	  & k \{k$^{\rm h}$\} [{\xng}] \\
Affricate	& 		& ts \{ts$^{\rm h}$\}	& t{\alvpalfric} \{t{\alvpalfric}$^{\rm h}$\}	& t{\retfric} \{t{\retfric}$^{\rm h}$\}	& 	  &	\\
Fricative	& f		& s	 	& {\alvpalfric}		& {\retfric} ({\xrUS})	& 	  & x	\\
Approximant	&		& 		& 		& 		& l 	  & \\
\hline
\end{tabular}
\end{center}
\caption
{ Feature values of consonants in Mandarin. Columns have different
Place classes and rows different Manner classes.  Aspirated consonants
are in braces \{\}, voiced in parentheses () and nasalized 
in square brackets [].  Note that {\xrUS} is a voiced fricative in
Mandarin, not an approximant.  w and j are absent as they were
treated as vowels.  }
\label{mandarin_featvals}
\end{table}

When comparing different languages, one often finds claims such as
``language X makes more use of such-and-such-a-contrast than language
Y''.
Quantifying FL allows one to answer several questions harder than `Does Xhosa
make more use of clicks than French?' The most detailed questions, of
course, require computations to be even more robust than they are at
the moment.

This section has computations of FL for Dutch, English and German from
CELEX \cite{celex} and for Mandarin based on the TDT3 Multilanguage
Text Version 2.0 corpus of transcriptions of Voice of America Mandarin
broadcasts.  In all cases calculations were based on word-frequency
pairs, with citation form pronunciations for the former and
frequencies from mostly written corpora.  The Mandarin word for VOA
was excluded from the word-frequency pairs.

Each syllable in the three European languages has a stress
component. $\Phi_{\tt str}$ = \{primary, secondary, unstressed\} for
English and \{present, absent\} for German and Dutch. Syllable stress
information was not available for Mandarin in our corpus, though of
course tonal information was. Therefore Mandarin syllables had just two
components, of type {\tt string<phn>} and {\tt ton}. 

Some of our calculations will involve distinctive features for
consonants.  We use the distinctive features Place, Manner, Nasality,
Voicing (for Dutch, English and German) and Aspiration (for Mandarin). All but the first two
are binary features. We arrange the features in a hierarchical scheme
that is a much simplified version of that proposed by
\namecite{oldfogey}. Features do not have to be specified for each
phoneme, e.g. Nasality is only specified for stops.  Table
\ref{mandarin_featvals} shows our arrangement of Mandarin features
while Table \ref{DEG_featvals} shows that for English, Dutch and
German. Note the following in the latter :

\begin{table}[h]
\begin{center}
\begin{tabular}{|l|c|c|c|c|c|c|c|c|c|}
\hline
		& Labial			& Den	& Alveolar	& P-A	& Lat	 & Pal		& Velar		& Uvu	& Glo \\
\hline
Approx.		& 		   {\it v}	&		& r		& 		& l	  & j		& w		& 		& 	\\
Fricative	& 		 f (v)	 	& {\xth} (\xdh)	& s (z)		& {\xsh} (\xzh)	& 	  & \Gch	& x	(\Dg)   &      (\Gr)	&  h 	\\
Affricate	&	pf			& 		& ts		& \xch (\xdzh)  &	  &		&		&		&	\\
Stop		& p (b) [m]	 		& 		& t (d) [n]	& 		& 	  &		& k (g)	[\xng]	&		& 	\\
\hline
\end{tabular}
\end{center}
\caption
{Feature values of consonants in Dutch, German and English that are
used in CELEX.  Columns have different Place classes and rows
different Manner classes. P-A stands for Post-Alveolar, Den for Dental, Lat for Laterals, Pal for
Palatals, Uvu for uvular and Glo for Glottal.}
\label{DEG_featvals}
\end{table}

\begin{itemize}
\item The exact place of several phonemes is dialect dependent, e.g. [r] and [x] in Dutch.
\item The dentals [{\xth}] and [{\xdh}] are present in English only.
\item The rhotic [r] is in English and Dutch only, [{\Gr}] in German only.
\item Dutch does not have the velar approximant [w], but instead the labial one [{\it v}]. 
\item Only Dutch has phoneme [x].
\item Only some borrowed words in Dutch have [g]. 
\item The palatal [{\Gch}]  occurs in only German and some borrowed English words. 
\item The affricates [pf] and [ts] are only found in German. 
\item The affricate [{\xch}] is not found in Dutch. 
\item CELEX does not code for a voiceless uvular fricative in Dutch or German, though the IPA does \cite{ipabook}.
\end{itemize}

\begin{table}
\begin{center}
\begin{tabular}{|l|l|r|r|}
\hline
Feature & Partition (non-singleton classes)& Syllables & Words \\
\hline
\textbf{ Aspiration } &&& \\
Mandarin      	& p$^{\rm h}$p t$^{\rm h}$t ts$^{\rm h}$.ts t{\alvpalfric}$^{\rm h}$.t{\alvpalfric} t{\retfric}$^{\rm h}$.t{\retfric} k$^{\rm h}$k     			&  16.7         &  2.7  \\
\hline
\textbf{ Voicing } &&& \\
Dutch         	& pb fv td sz {\xsh}{\xzh} kg x\Dg        			&  30.2         &  3.1  \\
English	        & pb fv {\xth}{\xdh} td sz {\xch}{\xdzh} {\xsh}{\xzh} kg       		&  23.3         &  4.5  \\
German		& pb fv td sz {\xch}{\xdzh} {\xsh}{\xzh} kg          		&  20.7         &  1.1  \\
  &&& \\  	
\hline 
\textbf{ Place } &&& \\  	
Dutch		& wlj fs{\xsh}hx {\Dg}vz{\xzh} ptk bdg mn{\xng}    		&  67.1         & 11.4  \\
English		& rljw f{\xth}s{\xsh}{\Gch}h v{\xdh}z{\xzh} ptk bdg mn{\xng}          	&  72.5         & 20.1  \\
German		& ljw fs{\xsh}{\Gch}h vz{\xzh} ptk bdg mn{\xng} {\xch}.pf.ts          	&  60.5         & 12.6  \\
Mandarin    	& ptk p$^{\rm h}$t$^{\rm h}$k$^{\rm h}$ mn{\xng} ts.t{\alvpalfric}.t{\retfric} ts$^{\rm h}$.t{\alvpalfric}$^{\rm h}$.t{\retfric}$^{\rm h}$  fs{\alvpalfric}x{\retfric}    			&  65.0         & 14.2  \\

  &&& \\  	
\hline
\textbf{ Manner } &&& \\
Dutch  		& wfp bv st dz sh \xzh{\xdzh} xk g{\Dg}       		&  27.1         &  4.5  \\
English 	& fp bv rst dz \xsh\xch \xzh{\xdzh} wk j{\Gch}  		&  39.2         & 11.4  \\
German		& fp.pf bv st.ts dz \xsh{\xch} \xzh{\xdzh} wk j{\Gch}     		&  27.4         &  8.0  \\
Mandarin   	& fp t.ts.s t{\alvpalfric}.{\alvpalfric} t{\retfric}.{\retfric} kx       			&  33.7         & 6.4  \\
  &&& \\  	
\hline
\textbf{ Nasality } &&&\\
Dutch         	& bm dn g\xng      				&  15.2         &  1.5  \\
English         & bm dn g\xng      				&  11.6         &  3.3  \\
German          & bm dn g\xng      				&  15.5         &  1.8  \\
Mandarin        & pm tn k\xng      				&   8.0         &  3.1  \\
  &&& \\  	
\hline
\textbf{ Tone } &&&\\
Mandarin    	& High.Rising.Low.Falling.Absent		&  107.5        & 21.3  \\
  &&& \\  	
\hline
\textbf{ Stress } &&& \\
Dutch   	& Present.Absent  				&  25.7         &  0.7  \\
English	  	& Primary.Secondary.Absent		   	&  26.9         &  0.1  \\
German  	& Present.Absent 				&  34.2         &  0.2  \\
\hline
\end{tabular}
\end{center}
\caption
{
Functional Load of several distinctive features in four languages. The
second column describes the non-singleton classes in the partition
used to obtain the FL value for a particular distinctive feature in a
language.  All values should be multiplied by 0.001. Phonemes represented
by more than one character are separated from others using a period, e.g. the
first Manner class for German has three phonemes : [p], [f] and [pf].
}
\label{FL_features_difflangs}
\end{table}

Table \ref{FL_features_difflangs} has FL values for the features defined above, while 
Table \ref{FL_groups_difflangs} has FL values for several sets of phonemes. 
The following conclusions can be drawn :

\begin{itemize}
\item Tones in Mandarin carry far more information than Stress in the
non-tonal languages. When word information is added, the FL of Stress
in the latter drops to almost nothing, while that for Mandarin remains
very high, having a far larger FL than Manner or Place.  In fact, as
shown in Table \ref{FL_groups_difflangs}, the FL of tone in Mandarin
is comparable to that of vowels (see \namecite{surgin_tones} for more
details).  This emphasizes the lexical role Tone plays in Mandarin, a
role clearly not played by Stress in the non-tonal languages.

\item Consonants have a higher FL than vowels. 

\item With respect to the way we have organized distinctive features,
Place has a higher FL than Manner.  However, consider also the more
specific case of alveolars and fricatives. The former have a very high
FL in English (as noticed in \namecite{pisoni}), Dutch and German,
over twice as high as that of fricatives despite the similar number of
phonemes in the two sets. But distinguishing between alveolars
involves working out Manner while distinguishing between fricatives
involves Place.

\item $FL_{\tt wrd}$ is always lower than $FL_{\tt syl}$. This is to
be expected, since knowledge of words and word boundaries is
additional information available to the listener that can be used to
make up for deficiencies elsewhere. 

\item All four languages place comparable amounts of FL on Place,
Manner and Nasality. Whether there is anything universal about this
remains to be seen. There certainly does not appear to be any
universal along the lines of stops having a higher/lower FL than
fricatives. On a side note, the latter values may be useful tools when
studying lenition in historical linguistics.

\item Mandarin makes far more use of affricate oppositions than German
or English.

\end{itemize}


\begin{table}
\begin{center}
\begin{tabular}{|l|l|r|r|}
\hline
Phoneme set		& Partition			  		& Syllables	& Words \\
\hline

\textbf{ Vowels } &&& \\
Dutch  	& &	125.5  & 51.5 \\
English &  & 133.0        & 48.5 \\
German &   & 161.3	 & 42.2 \\
Mandarin   & &  91.0	& 22.1 \\
\hline

\textbf{ Consonants   }  & && \\
Dutch	&  &335.8        & 192.5 \\
English &  &309.8        & 176.4 \\
German  &  &335.6	& 153.8 \\
Mandarin&  &234.7        & 80.5 \\
\hline

\textbf{ Labials } &&& \\         
Dutch		& pbmfvw        				&  36.5         &  8.7 \\
English         & pbmfv         &  25.2         &  5.9 \\
German & pbmfv.pf        &  23.0         &  3.6 \\
Mandarin & p$^{\rm h}$pfm & 10.0         &  1.8 \\
\hline

\textbf{ Alveolars } &&& \\
Dutch       & tdsznlr       &  101.5        & 37.5 \\
English & tdsznrl &   98.2         & 41.5 \\
German & tdsznl.ts       &  89.3         & 22.7 \\
Mandarin & t$^{\rm h}$t.ts$^{\rm h}$.ts.sn &   24.7         & 7.5 \\
\hline

\textbf{ Velars } &&& \\
Dutch &   kg{\xng}x\Dg &  20.6         &  0.8 \\
English & kg{\xng}w  &   6.7         &  1.3 \\
German & kg{\xng}w          &   5.5         &  0.1 \\
Mandarin & k$^{\rm h}$kx{\xng}          &   8.8         &  1.4 \\
\hline

\textbf{ Nasals } &&& \\
Dutch  & mn{\xng}   &  12.0         &  2.0 \\
English  & mn{\xng}   &  11.5         &  2.8 \\
German & mn{\xng} & 14.4         &  4.4 \\
Mandarin & mn{\xng}   &  16.2         &  3.1 \\
\hline

\textbf{ Fricatives } &&& \\
Dutch      & fvrsz{\xsh}{\xzh}xh     &  39.1         &  7.8 \\
English       & fv{\xth}{\xdh}sz{\xsh}{\xzh}{\Gch}h    &  39.6         & 17.8 \\
German & fvrsz{\xsh}{\xzh}{\Gch}h     &  53.2         & 14.1 \\
Mandarin & fs{\alvpalfric}{\xrUS}x{\retfric}        &  20.7         &  5.1 \\
\hline

\textbf{ Affricates } &&& \\
English & {\xch}{\xdzh}    &   0.8         &  0.1 \\
German & {\xch}{\xdzh}.pf.ts          		&   0.7         &  0.0 \\
Mandarin & ts$^{\rm h}$.ts.t{\alvpalfric}.t{\alvpalfric}$^{\rm h}$.t{\retfric}.t{\retfric}$^{\rm h}$        	&  25.1         &  5.1 \\
\hline

\textbf{ Stops } &&& \\
Dutch    & ptkbdg        &  56.3         & 10.8 \\
English & ptkbdg &  43.3         & 10.6 \\
German &   ptkbdg      &  50.1         &  4.5 \\
Mandarin &      p$^{\rm h}$t$^{\rm h}$k$^{\rm h}$ptk  &  29.3         &  6.2 \\ 
\hline
\end{tabular}
\end{center}
\caption{
The FL of  several sets of phonemes in four languages. The second column describes the
non-singleton classes in the partition corresponding to each set and language.
All values should be multiplied by 0.001. }
\label{FL_groups_difflangs}
\end{table}


\section{An application in historical linguistics}\label{applic_histling}

Suppose we wish to investigate Martinet's hypothesis \cite{martinet}
that FL plays some role in phoneme mergers. To do this properly,
several examples of mergers are necessary, with appropriate corpora
for each case. This is hard to get. However, we do have one example
that we can use to illustrate the method of investigation.

As described by \namecite{zee}, [n] has merged with [l] in Cantonese 
in word-initial position in the last fifty years.
We used a
word-frequency list derived from CANCORP \cite{cancorp}, a corpus of
Cantonese child-adult speech which has conveniently coded [n] and [l]
as they would have occurred before the merger. Merging only in
word-initial position, we computed $FL_{\tt wrd}$(n,l), which is a
completely meaningless value by itself. We therefore also computed
$FL_{\tt wrd}$(x,y) for all consonants in Cantonese, and found that
$FL_{\tt wrd}$(n,l) was larger than over 70\% of them.   That tells us 
that the [n]-[l] contrast did have a high FL before the merger. 


\begin{table}[h]
\begin{center}
\begin{tabular}{|l|rrrrrrrrr|}
\hline
$x$                          & l  & p$^{\rm h}$  & t$^{\rm h}$  & k$^{\rm h}$  & p  & t & k  & w  & ts  \\
$FL_{\tt wrd}($n$,x)$    & 9.0& 2.8& 0.7& 3.4& 0.1& 1.4& 7.0& 0.4& 0.3 \\ 
\hline
$x$                          & ts$^{\rm h}$    & m   & h   & f   & s   & {\xng}  & k$^{\rm hw}$   &  k$^{\rm w}$  & j  \\
$FL_{\tt wrd}($n$,x)$    & 4.8 & 9.1 & 2.5 & 2.3 & 2.2 & 1.1 & 0   & 0.0 & 3.7 \\
\hline
\end{tabular}
\end{center}
\caption{
Functional load values of the opposition of [n] with other consonants in Cantonese before it merged with [l] in word-initial position.
Values  computed with the CANCORP corpus, $n=1$ and ${\tt T}={\tt wrd}$. 
Values should be multiplied by $10^{-4}$. 
}
\label{cantpremerger}
\end{table}

Table \ref{cantpremerger} shows $FL_{\tt wrd}($n$,x)$ for all
word-initial consonants $x$.  The results are clear, and rather
startling. Of all the consonants [n] could have merged with, it merged
with the second `worst' (in an optimal sense) choice! This result adds
weight to those of \namecite{king}, the only previous corpora-based
test of Martinet's hypothesis.

\section{An application in child language acquisition}\label{applic_acq}

As mentioned early in the paper, there has been a need in this field
for a comprehensive FL measure for some time. A major question is what
factors affect the age at which children acquire sounds in the
language. This has been investigated recently by \namecite{stokes} for
consonants in three languages. 

The frequency of a sound is not a consistent (across languages)
predictor of when a child start to use it. For example, they find that
frequency correlates very significantly with age of acquisition in
Cantonese children, but the corresponding correlation for English is
not significant at all. In fact, the most common consonant in English
speech is /\xdh/, which is among the last children acquire.

On the other hand, the frequency of a phoneme is not the only measure
of its importance to the language. One can estimate the FL of a
phoneme as well, as described in Section \ref{FLsinglephoneme}. Recall
that $FL(x) = \sum_{y\in S(x)-x} P(x,y) FL(x,y)$, where $S(x)$ is the
set of `similar' phonemes to $x$, and $P(x,y)$ is the probability that
$x$ merges with $y$.

\namecite{stokes} find that when $x$ is a consonant, if $S(x)$ is
taken to be the set of consonants with the same place and laryngeal
setting, and $P(x,y)$ is proportional to the frequency of $y$, then
the FL of a phoneme is significantly correlated ($p<0.05$) to age of
acquisition in the three languages they check, namely Cantonese,
English and Mandarin. This makes a lot of sense if children find if
easier to get place and laryngeal setting (voicing, aspiration) right
than manner. Note that age of acquisition refers to initial appearance
of a sound in the child's phonetic inventory, not how the child uses
it in its phonemic system after that.


\section{Applications in automatic speech recognition}\label{applic_ASR}

FL has, of course, already been used in the ASR community by
\namecite{carter}; the work of \namecite{shipman},
\namecite{huttenlocher} and \namecite{kassel} should also be
mentioned. 

That syllables in English can be represented as a sequence of phonemes
plus a stress component, the cost of whose removal can be computed, is
nothing new. Extending this to tonal languages in the natural way is a
simple step, but it has not been, to our knowledge, been taken before,
and has already produced (see \namecite{surgin_tones}) the important
result that an ASR system for Mandarin that does not try to identify the underlying
tone of a syllable can only work as well as one that does identify tone
but does not identify vowels! Rephrasing PIE as FL might sound superficial; but even
if rephrasing does not result in additional answering power, it can
result in additional question-asking power.

In any case, our FL framework is an extension rather than a simple
rephrasing. For example, detailed analyses of a phonetically-based ASR
system can throw up problems that it would be useful to know the
importance of --- if they are not important, they can be ignored.
Suppose an ASR system often errs in deciding whether there is or is
not a [j] before a high vowel. A decision is taken to always ignore
the presence of such a [j] (or alternatively, to impose its presence
even when absent) --- how much information will be lost by doing so?
By finding the FL of such a contrast, which is represented by the rule
below, researchers can make a better informed decision.

$$
g_{{\tt syl},\theta}(\langle p_1\ldots p_m,s\rangle) = 
\left\{
\begin{array}{ccl}
\langle p_1\ldots p_{i-1}p_{i+1}\ldots p_m,s\rangle & {\rm if } &
p_i={\rm [y]\ \& \ }  p_{i+1}\in\{\rm high\ vowels\} \\
\langle p_1\ldots p_m,s\rangle & {\rm if } & {\rm not} 
\end{array}
\right.
$$

\section{Interpreting FL values}\label{interpret}

A serious-looking limitation of FL values is that they are
relative rather than absolute. However, this still allows them to be
used in several applications. One example is correlation analysis,
since $corr(X,Y)=corr(aX,Y)$ and $corr(\log(X),\log(Y)) =
corr(\log(aX),\log(Y))$ for any $a>0$. So if we want to see if
there is any correlation between FL, or log FL, and some other
parameter, we can do so with relative FL values. 

Another way to interpret FL values is comparing them with other FL
values computed the same way. For example, in Section \ref{difflang}
we wanted to see how important tones were in Mandarin, and got some
number for FL(tones). Knowing the importance of identifying vowels, we
compared FL(vowels) with FL(tones). The closeness of the values showed
that tones were at least as important as vowels in Mandarin.

\section{Conclusion}

A language makes use of contrasts to convey information; we have
proposed and empirically tested a framework for measuring the amount
of use. Further statistical tests and improvements of the measure are  required, but we
believe several linguistic questions can already be moved from the realm of description and speculation to testable
hypotheses.


\section*{Acknowledgements}

We are very grateful to Gina-Anne Levow for help with the Mandarin
data and several very useful discussions, Stephanie Stokes for the
Cantonese data and introducing us to the child language literature,
Bert Peeters for explaining to us how FL is viewed in the Martinet
tradition, Yi Xu for details of the behaviour of tones in Mandarin,
and John Goldsmith for several suggestions regarding the readability
of this paper.  Thanks also
go to Sean Fulop, Derrick Higgins, Jinyun Ke, Caroline Lyon and Howard
Nusbaum for their comments on earlier versions of this paper.


\bibliographystyle{fullname}
\bibliography{surendran-03}

\begin{thebibliography}{}

\bibitem[\protect\citename{Albro}1993]{amar}
Albro, Daniel M.
\newblock 1993.
\newblock {\em ``AMAR, a Computational Model of Autosegmental Phonology''}
\newblock MIT Technical Report AITR-1450, Cambridge, MA.

\bibitem[\protect\citename{Baayen, Piepenbrock and Gulikers}1995]{celex}
Baayen, R.~H., Piepenbrock, R., and Gulikers, L., 
\newblock 1995.
\newblock {\em The Celex Lexical Database} (Release 2).
\newblock Linguistic Data Consortium, 	Univ.~of Pennsylvania (Distributor), Philadelphia, PA.


\bibitem[\protect\citename{Carter}1987]{carter}
Carter, David M.
\newblock 1987.
\newblock An information-theoretical analysis of phonetic dictionary access.
\newblock {\em Computer Speech and Language}  2:1--11.

\bibitem[\protect\citename{Chao}1968]{chao}
Chao, Y.~R.
\newblock 1968.
\newblock {\em A Grammar of Spoken Chinese}
\newblock University of California Press, Berkeley.

\bibitem[\protect\citename{Greenberg}1959]{greenberg}
Greenberg, H.~H.
\newblock 1959.
\newblock A method of measuring functional yield as applied to tone in
African languages.
\newblock {\em Georgetown University Monograph Series on Language and Linguistics} 12:7--16.

\bibitem[\protect\citename{Godfrey, Holliman and McDaniel}1992]{swb}
Godfrey, J., Holliman E. and McDaniel, J.
\newblock 1992.
\newblock  Telephone speech corpus for research and development.
\newblock {\em Proc. IEEE ICASSP}, pp. 517--520.

\bibitem[\protect\citename{Goldsmith}1976]{goldsmith}
Goldsmith, John.
\newblock 1976.
\newblock {\em Autosegmental Phonology.}
\newblock PhD Thesis,  Department of Linguistics, Massachusetts Institute of Technology.

\bibitem[\protect\citename{Hockett}1955]{hockett}
Hockett, Charles F.
\newblock 1955.
\newblock {\em A Manual of Phonology}.
\newblock International Journal of American Linguistics 21(4), Indiana University Publications.

\bibitem[\protect\citename{Hockett}1967]{hockett67}
Hockett, Charles F.
\newblock 1967.
\newblock The quantification of functional load.
\newblock {\em Word} 23:320--339.

\bibitem[\protect\citename{Huttenlocher}1985]{huttenlocher}
Huttenlocher, D. 
\newblock Exploiting sequential phonotactic constraints in recognizing
spoken words.
\newblock {\em MIT AI Lab Memo} 867.

\bibitem[\protect\citename{Ingram}1989]{ingram}
Ingram, David.
\newblock 1989.
\newblock {\em First language acquisition: method, description and
explanation}.
\newblock Cambridge University Press, Cambridge, UK.

\bibitem[\protect\citename{IPA Handbook}1999]{ipabook}
International Phonetic Association
\newblock 1999.
\newblock {\it Handbook of the International Phonetic Association}
\newblock Cambridge University Press, Cambridge, UK.

\bibitem[\protect\citename{Kassel}1990]{kassel}
Kassel, Robert.
\newblock 1990.
\newblock {\em ``An informational-theoretical approach to studying
phoneme collocational constraints''}
\newblock MS Thesis, EECS Department, MIT. 

\bibitem[\protect\citename{King}1967]{king}
King, Robert D.
\newblock 1967.
\newblock Functional load and sound change.
\newblock {\em Language}, 43:831--852. 

\bibitem[\protect\citename{Kontoyannis}1997]{kontoyannis} 
Kontoyannis, I.
\newblock 1997.
\newblock {\em ``The complexity and entropy of literary styles''}
\newblock NSF Technical Report No. 97, Department of Statistics, Stanford University.

\bibitem[\protect\citename{Ku\v{c}era}1963]{kucera}
Ku\v{c}era, Henry.
\newblock 1963.
\newblock Entropy, redundancy and functional load.
\newblock {\em American Contributions to the Fifth International Conference
of Slavists (Sofia)}: 191--219.


\bibitem[\protect\citename{Ladefoged}1997]{oldfogey}
Ladefoged, Peter.
\newblock 1997.
\newblock Linguistic phonetic descriptions.
\newblock Chapter 19 in {\it The Handbook of Phonetic Sciences}
\newblock Hardcastle and Laver (eds.), Blackwell Publishers.

\bibitem[\protect\citename{Lass}1980]{lass80}
Lass, Roger.
\newblock 1980.
\newblock {\em On Explaining Language Change}.
\newblock Cambridge University Press.

\bibitem[\protect\citename{Lass}1997]{lass97}
Lass, Roger.
\newblock 1997.
\newblock {\em Historical Linguistics and Language Change}.
\newblock Cambridge University Press.

\bibitem[\protect\citename{Lee et al}1996]{cancorp}
Lee,~T.H.T., Wong,~C.H., Leung,~C.S., Man,~P., Cheung,~A., Szeto,~K. and Wong,~C.S.P. 
\newblock 1996.
\newblock {\em The development of grammatical competence in Cantonese-speaking children.}
\newblock Report of a project funded by Research Grants Council, Chinese University of Hong Kong.

\bibitem[\protect\citename{Martinet}1955]{martinet}
Martinet, Andr\'e.
\newblock 1955.
\newblock {\em \'Economie des Changements Phon\'etiques.}
\newblock Bern, Francke.


\bibitem[\protect\citename{Mathesius}1929]{mathesius}
Mathesius, Vil\'em.
\newblock 1929.
\newblock La structure phonologique du lexique du tch\`eque moderne.
\newblock {\em Travaux du Cercle Linguistique de Prague}, 1:67-84.


\bibitem[\protect\citename{Meyerstein}1970]{meyerstein}
Meyerstein, R.~S.
\newblock 1970.
\newblock  Functional load: descriptive limitations, alternatives
of assessment and extensions of application
\newblock {\em Janua Linguarum, Series Minor} \#99.


\bibitem[\protect\citename{Peeters}1992]{peeters}
Peeters, Bert.
\newblock 1992.
\newblock {\em Diachronie, Phonologie et Linguistique Fonctionnelle}.
\newblock Louvain-la-Neuve, Peeters.


\bibitem[\protect\citename{Pisoni et al}1985]{pisoni}
Pisoni, D.B., Nusbaum, H.C., Luce, P.A. and Slowiaczek, L.M.
\newblock 1985.
\newblock Speech perception, word recognition and the structure of the
lexicon
\newblock {\em Speech Communication} 4: 75--95.

\bibitem[\protect\citename{Pye, Ingram and List}1987]{pye}
Pye, Clifton, Ingram, David and List, Helen.
\newblock 1987.
\newblock A comparison of initial and final consonant acquisition in
English and Quich\'e.
\newblock in K.~E.~Nelson and A.~van Kleek (eds.), {\em Children's
language} Vol. 6. 
\newblock Erlbaum, Hillsdale, NJ.

\bibitem[\protect\citename{Shannon}1951]{shannon}
Shannon, Claude E.
\newblock 1951.
\newblock Prediction and entropy of printed English.
\newblock {\em Bell Systems Technical Journal} 30:50-64.

\bibitem[\protect\citename{Shipman and Zue}1982]{shipman}
Shipman, David W. and Zue, Victor W.
\newblock 1982.
\newblock Properties of large lexicons; implications for advanced
isolated word recognition systems.
\newblock {\em Proc. IEEE ICASSP} 546--549.

\bibitem[\protect\citename{So and Dodd}1995]{sododd}
So, Lydia K.~H., and Dodd, Barbara J.
\newblock 1995.
\newblock The acquisition of phonology by Cantonese-speaking children.
\newblock {\em J.~Child Lang.} 22: 473-495.

\bibitem[\protect\citename{Stokes and Surendran}2003]{stokes}
Stokes, Stephanie and Surendran, Dinoj.
\newblock 2003.
\newblock Articulatory complexity, ambient frequency and functional
load as predictors of consonant development in children.
\newblock {\em Submitted.} 

\bibitem[\protect\citename{Surendran and Niyogi}2003]{surniy_evol}
Surendran, Dinoj and Niyogi, Partha. 
\newblock 2003.
\newblock Questioning the role of communicative efficiency in language evolution.
\newblock {\em To be submitted.} 

\bibitem[\protect\citename{Surendran and Levow}2003]{surgin_tones}
Surendran, Dinoj and Levow, Gina-Anne.
\newblock 2003.
\newblock The functional load of tone in Mandarin is as high as that of vowels.
\newblock {\em Submitted.} 

\bibitem[\protect\citename{Trubetzkoy}1939]{trubetzkoy}
Trubetzkoy, Nikolay.
\newblock 1939.
\newblock {\em Grundz\"{u}ge der phonologie}.
\newblock Travaux du Cercle Linguistique de Prague 7.

\bibitem[\protect\citename{Wang}1967]{wang}
Wang, William S-Y.
\newblock 1967.
\newblock The measurement of functional load.
\newblock {\em Phonetica} 16:36--54.

\bibitem[\protect\citename{Xu}1993]{xudiss}
Xu, Yi.
\newblock 1993.
\newblock {\em Contextual tonal variation in Mandarin Chinese}.
\newblock PhD Thesis, Department of Linguistics, The University of Connecticut.

\bibitem[\protect\citename{Xu}1994]{xucoartictone}
Xu, Yi.
\newblock 1994.
\newblock Production and perception of coarticulated tones.
\newblock {\em J.~Acoust.~Soc.~Am.} 95: 2240-2253.

\bibitem[\protect\citename{Zee}1999]{zee}
Zee, Eric.
\newblock 1999.
\newblock Change and variation in the syllable-initial and
syllable-final consonants in Hong Kong Cantonese. 
\newblock {\em Journal of Chinese Linguistics} 27, 120--167. 


\end{thebibliography}

\end{document}